\documentclass[10pt,twocolumn,letterpaper]{article}

\usepackage[pagenumbers]{cvpr}
\usepackage[T1]{fontenc}
\usepackage{microtype}
\usepackage{placeins}
\usepackage{array}
\usepackage{multirow}
\definecolor{cvprblue}{rgb}{0.21,0.49,0.74}
\usepackage[pagebackref,breaklinks,colorlinks,allcolors=cvprblue]{hyperref}
\hypersetup{pdftitle={ReactWorld: Structuring Interaction Responses for Embodied Video Generation},pdfauthor={Anonymous},pdflang={en-US}}

\title{ReactWorld: Structuring Interaction Responses\\
	for Embodied Video Generation}

\author{
	Hanwei Liu$^{1}$ \quad
	Huijia Zhang$^{1}$ \quad
	Hanzhang Wang$^{1}$ \quad
	Guangyu Liu$^{2}$\\
	Hanbei Zhan$^{3}$ \quad
	Linjie Hu$^{1}$ \quad
	Bin Xiao$^{1}$ \quad
	Xuan Zhou$^{1}$ \quad
	Dong Zhang$^{1}$\\[4pt]
	{\normalsize $^{1}$Huawei Technologies Co., Ltd.}\\
	{\normalsize $^{2}$Tongji University \quad
		$^{3}$Zhejiang University}\\[4pt]
	{\small \url{https://metaengine-world.github.io/ReactWorld/}}
}

\begin{document}
	
	\twocolumn[{
		\renewcommand\twocolumn[1][]{#1}
		\maketitle
		\begin{center}
			\includegraphics[width=.98\textwidth]{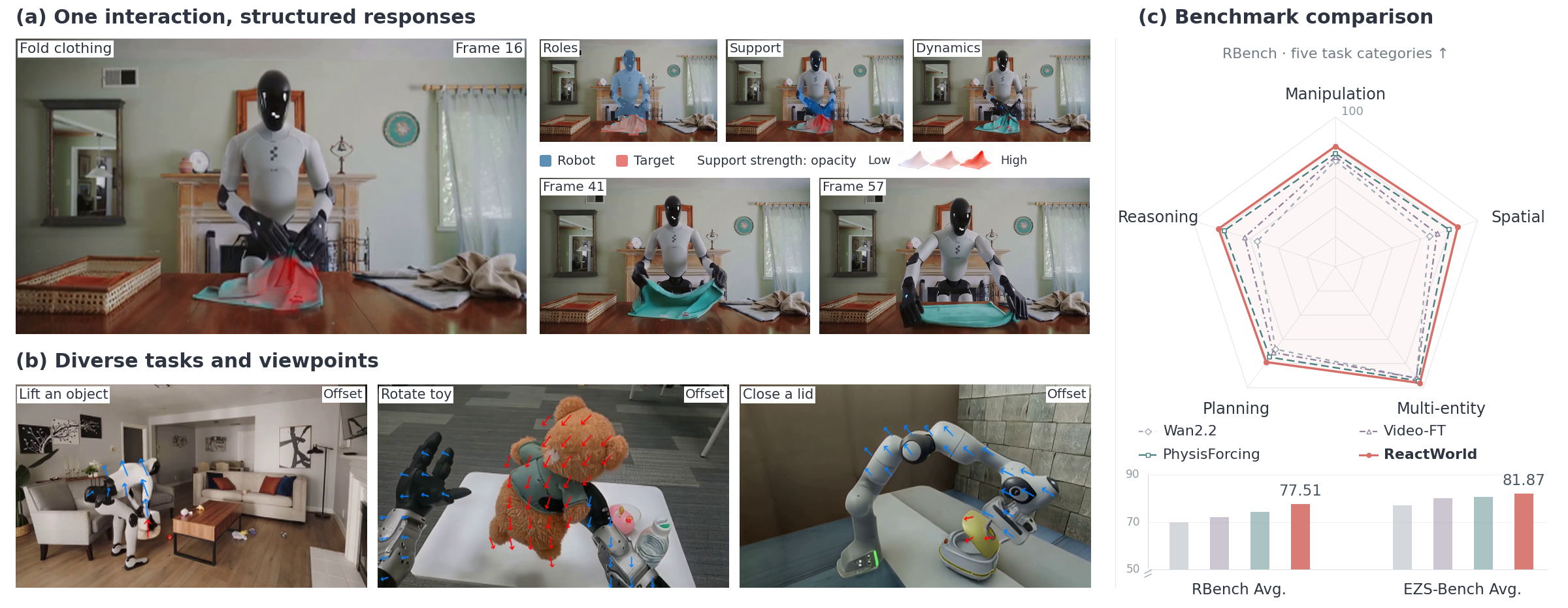}
			\captionsetup{type=figure}
			\caption{\textbf{Structured interaction responses and generation performance.} (a) Roles, local motion support, interval dynamics, and subsequent states in garment folding. Overlay opacity encodes support strength, and arrows denote 2D displacement. (b) Robot and target displacements across tasks and viewpoints. (c) RBench scores on five task categories, and average scores on RBench and EZS-Bench.}
			\label{fig:teaser}
		\end{center}
	}]
	
\begin{abstract}
Embodied video world models predict scene changes caused by robot actions, yet generated object responses may remain inconsistent with task intent and contact dynamics. Critical responses in embodied scenes often concentrate in local regions and specific time intervals. Inspired by the use of explicit intermediate steps in chain-of-thought reasoning, we introduce ReactWorld, an embodied video generation model guided by structured interaction anticipation. In the generation backbone's feature space, an interaction response field predictor anticipates how the interaction will unfold, explicitly describing participant roles, future motion response intensity, and local motion changes. A response feedback adapter forms increments to object features from the difference between the object branch's outputs with predicted and zeroed local dynamics inputs. It also converts predicted displacements into positional biases for temporal attention. These two mechanisms guide object feature updates and information retrieval across frames, respectively. Through interaction anticipation and generative feedback, ReactWorld achieves an average score of 77.51 on RBench, outperforming PhysisForcing by 3.29 points with improvements in all nine categories. On EZS-Bench, it achieves a Domain score of 86.98 and an average score of 81.87.
\end{abstract}

\section{Introduction}
\label{sec:intro}
Embodied video world models represent the consequences of robot actions as future visual sequences, allowing interactions to be previewed~\cite{ref11_robodreamer,ref17_irasim}. Unlike general video generation, embodied interactions require object changes to agree with task intent, contact relations, and motion dynamics~\cite{ref03_physisforcing,ref02_impact}. Image-to-video models can generate realistic appearance and coherent motion, and exhibit the ability to capture some physical regularities. However, visually plausible motion does not ensure that objects respond reliably to interaction conditions~\cite{wang2026gauge,chen2026viper}. A target may move before contact, a grasped object may detach from the gripper, or nearby objects may respond unnecessarily. Even when the overall video is smooth, such local errors can alter the outcome of grasping, placement, or tool use.

The dynamics that determine an interaction's success often occur at a few locations and critical moments; their importance is not proportional to their visual extent~\cite{ref03_physisforcing,ref02_impact}. When folding a garment, for example, the entire garment is the target, but the initial response may be confined to the corner being lifted. VideoJAM~\cite{ref01_videojam} identifies an imbalance between appearance fitting and motion learning, while IMPACT~\cite{ref02_impact} shows that globally averaged denoising supervision can underweight sparse regions containing dynamic objects. These findings suggest that embodied video generation must learn not only motion, but also which objects participate, where local responses occur, and when they unfold.

Existing methods strengthen dynamics modeling at different stages. PhysisForcing~\cite{ref03_physisforcing} aligns intermediate features with trajectories and relations between regions; Motion-I2V~\cite{ref04_motioni2v} and VideoJAM~\cite{ref01_videojam} use predicted motion for temporal feature propagation or generation guidance. VPT~\cite{ref05_vpt} introduces entity roles and uses modality-decoupled denoising and auxiliary loss annealing to reduce optimization conflicts and error propagation in joint modeling. Meanwhile, RecCAR~\cite{ref06_reccar} finds that bidirectional connections do not guarantee that auxiliary modalities constrain video generation equally effectively when jointly generating video with human motion and audio. These advances provide a foundation for motion modeling and generation guidance, but coherent motion alone does not ensure that object responses follow task intent and contact dynamics. For embodied interactions, the key is to specify where, when, and how objects should respond, and to make these task-relevant local dynamics effective conditions for generation.

\begin{figure*}[!t]
\centering
\includegraphics[width=\textwidth]{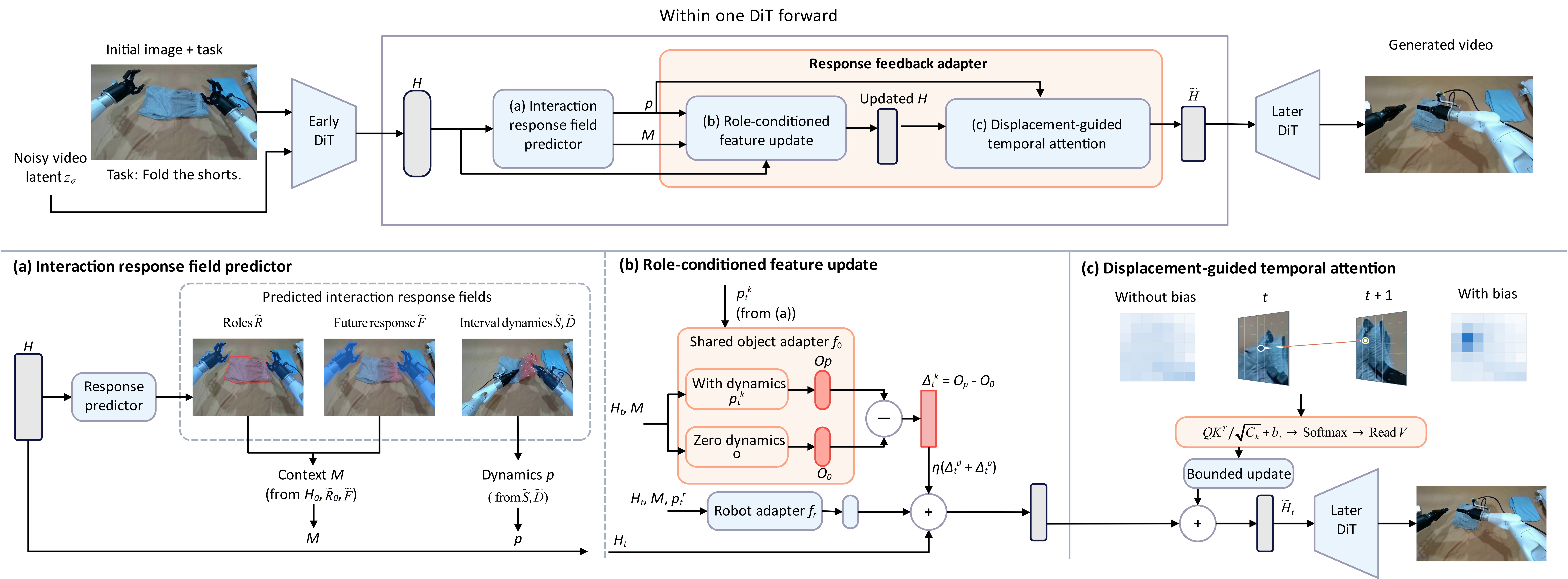}
\caption{\textbf{Overview of ReactWorld.} Top: response prediction and generative feedback within each denoising forward pass. Bottom: (a) predicting roles, future responses, and interval dynamics to construct interaction context and local dynamic conditions; (b) extracting object-response increments relative to a zero-dynamics reference and updating features by role; and (c) converting gated, clipped displacements into positional biases within a fixed candidate window. Biases are added to query--key logits before softmax and feature aggregation, followed by bounded residual feedback. Scene and response visualizations illustrate the mechanism.}
\label{fig:pipeline}
\end{figure*}

Inspired by chain-of-thought's use of explicit intermediate steps to structure complex reasoning~\cite{wei2022cot}, we propose ReactWorld, an embodied video generation model guided by structured interaction anticipation. Response prediction and generative feedback are performed within the same denoising forward pass, as shown in Fig.~\ref{fig:pipeline}.

Given an initial image and a task instruction, the \textbf{interaction response field predictor} uses intermediate backbone features to anticipate how the interaction will unfold: which roles objects play, which local regions will respond and how strongly, and how those regions will subsequently move.

The \textbf{response feedback adapter} pools interaction context using roles and future motion response intensity. Holding the features and context fixed, it takes the difference between the object branch's outputs with predicted and zeroed local dynamics inputs to form an increment to object features. It then converts predicted displacement into a temporal attention bias within a fixed local window in the next state. The former guides object feature updates; the latter guides where information is retrieved across frames. Inference requires only a task instruction and an initial image, without external role maps or trajectories.

ReactWorld achieves an average RBench score of 77.51, compared with 71.94 for Video-FT trained on the same data, and improves on PhysisForcing by 3.29 points. On EZS-Bench, its Domain and average scores reach 86.98 and 81.87. These results indicate that ReactWorld better captures the relationship between task requirements and object dynamics in embodied scenes, improving task completion and physical plausibility.

Our main contributions are:
\begin{itemize}
	\item We propose ReactWorld, an embodied video generation model guided by structured interaction anticipation, explicitly predicting participant roles, future motion response intensity, and interval dynamics in the backbone feature space.
	\item We develop an interaction response field predictor and a response feedback adapter, using output differences relative to a zero-dynamics reference to update object features and displacement-based attention biases to guide cross-frame aggregation.
	\item Improved overall scores on RBench and EZS-Bench, with more reliable interaction generation in terms of task completion, participant stability, and physical plausibility.
\end{itemize}

\section{Related Work}
\label{sec:related}
\paragraph{Embodied video world models.}
Video generation models~\cite{ref10_cosmos,ref11_robodreamer,ref12_genieenvisioner,ref13_gr2,ref16_unisim,ref17_irasim} offer new approaches to predicting the outcomes of robot actions and synthesizing experience. Gen2Act~\cite{ref08_gen2act} combines task-relevant human video generation with video-conditioned policies, while DreamGen~\cite{ref09_dreamgen} uses video world models to synthesize robot trajectories for downstream policy learning. These methods extend video generation to robot learning. ReactWorld focuses on object responses in generated videos, using task-relevant interaction anticipation to guide local dynamics.

\paragraph{Interaction and physical supervision.}
Prior work~\cite{ref02_impact,ref03_physisforcing,ref07_physalign,ref14_videorepa,ref15_roboscape} strengthens interaction modeling through motion, geometry, and regional relations. PhysisForcing~\cite{ref03_physisforcing} aligns video features with point trajectories and regional relations. PhysAlign~\cite{ref07_physalign} combines simulated geometric labels with relational feature alignment, while IMPACT~\cite{ref02_impact} reweights denoising supervision using object interaction information. ReactWorld also constructs interaction dynamics supervision from training videos, and uses its predictions as internal denoising conditions for object feature updates and temporal aggregation.

\paragraph{Motion modeling and generation guidance.}
MotionCtrl, DragNUWA, DragAnything, Tora, and Motion Prompting control video motion through poses, trajectories, or entity conditions~\cite{ref18_motionctrl,ref19_dragnuwa,ref20_draganything,ref21_tora,ref22_motionprompting}. Motion-I2V, VideoJAM, and Track4Gen use predicted motion or features learned with motion supervision for generation guidance and feedback~\cite{ref04_motioni2v,ref01_videojam,ref23_track4gen}. VPT introduces entity roles and addresses optimization challenges in multimodal joint modeling~\cite{ref05_vpt}. Building on motion guidance and feature feedback, ReactWorld organizes interaction roles, future motion response intensity, and interval motion as internal conditions. It updates object features using output differences relative to a zero-dynamics reference and guides cross-frame information retrieval with displacement-based attention biases in a fixed local window.

\section{Method}
\label{sec:method}
ReactWorld anticipates task-relevant interaction dynamics in the feature space of the generation backbone and feeds them back into video generation. It represents interaction requirements through participant roles, future local responses, and interval motion, then uses these predictions to guide object feature updates and information retrieval across frames. Fig.~\ref{fig:pipeline} provides an overview.

\subsection{Preliminaries}
Given an initial image $I_0$ and a task instruction $c$, the model progressively generates future video in latent space. Let $z$ denote the latent representation of a training video, $\epsilon$ Gaussian noise, and $\sigma$ the noise level. The noisy input is $z_\sigma=(1-\sigma)z+\sigma\epsilon$.

Let $u_\theta$ denote the backbone's velocity prediction network. Its output $\widehat u_\sigma=u_\theta(z_\sigma,I_0,c,\sigma)$ describes the rate of change of the video latent along the noise path. We train with the flow-matching objective:
\begin{equation}
\mathcal L_{\mathrm{video}}
=
\mathbb E\left[
\left\|
\widehat u_\sigma-(\epsilon-z)
\right\|_2^2
\right]
\label{eq:flowmatching}
\end{equation}

At each denoising step, we read intermediate features $\mathbf H$ from the generation backbone, where $\mathbf H_t(x)$ denotes the feature at temporal position $t$ and spatial position $x$ in the latent video sequence. The interaction response field predictor estimates interaction dynamics from these features. The response feedback adapter updates the features, and the remaining backbone layers continue video prediction.

\begin{figure*}[!t]
\centering
\includegraphics[width=\textwidth]{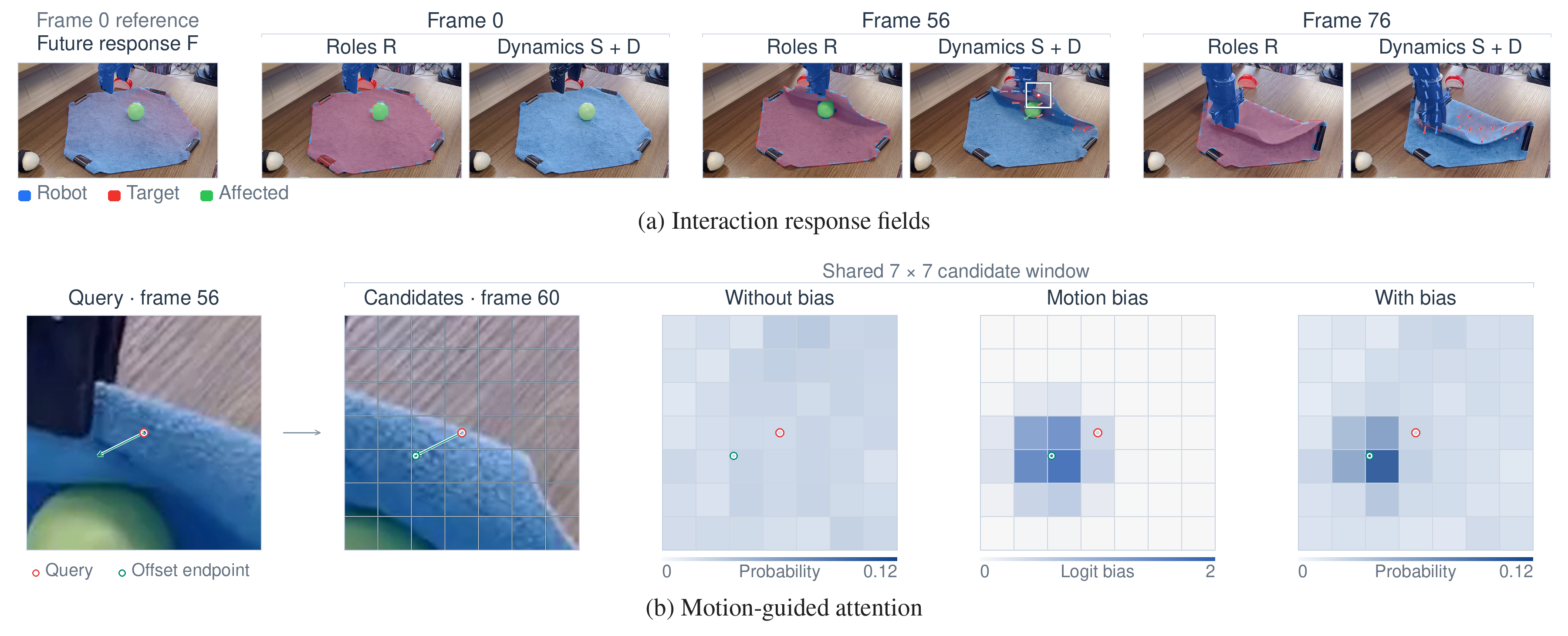}
\caption{\textbf{Structured response predictions and displacement biases.} (a) The future response field referenced to the first frame, together with object roles, motion support, and displacement at frames 0, 56, and 76, overlaid on the corresponding RGB frames. (b) The same query and a shared $7\times7$ candidate window for the interval from frame 56 to 60, showing attention without displacement bias, the displacement bias, and attention with bias. Red and green circles indicate the query and displacement endpoint, respectively.}
\label{fig:response}
\end{figure*}

\subsection{Interaction Response Field Predictor}
The task instruction specifies the interaction goal, while video generation must further characterize the participants, local responses, and motion over time. As shown in Fig.~\ref{fig:pipeline}(a), we organize this information into an interaction response field $\mathcal Y=(R,F,S,D)$, enabling explicit anticipation of the spatial distribution and evolution of interaction responses.

\paragraph{Participant roles $R$.}
Conditioned on the task instruction, the role map assigns each location to the robot $r$, direct manipulation target $d$, affected object $a$, or background $b$. In Fig.~\ref{fig:response}(a), the manipulated cloth is the direct target, while the ball on it is the affected object. These roles distinguish directly manipulated objects from those affected indirectly within the interaction context. We use $\mathcal K=\{r,d,a\}$ for the three interaction roles.

\paragraph{Future motion response intensity $F$.}
The role map describes the extent of participating objects, but different parts of an object need not respond equally. For example, when manipulating a garment, a lifted corner and a part still resting on the support surface may exhibit different response intensities. $F$ represents future motion response intensity for each role at each location in first-frame coordinates, where $F^k(x)\in[0,1]$ is the continuous response value for role $k$ at location $x$. This links local differences in subsequent motion to the initial scene, allowing the feedback module to emphasize features relevant to future responses within participating objects.

\paragraph{Interval motion $(S,D)$.}
While $F$ summarizes local responses over the full prediction horizon for each role, $S,D$ further characterize their motion regions and displacements at successive stages. Here, $S_t^k(x)\in[0,1]$ is a continuous weight indicating the motion region of role $k$ over interval $t\rightarrow t+1$, which we call motion support; $D_t^k(x)$ describes the motion direction and distance at that location. Both use coordinates at the start of the interval, with displacement measured in feature grid units, providing time-varying conditions for local dynamics feedback.

Four lightweight prediction heads read these response fields from intermediate video features:
\begin{equation}
\begin{aligned}
\widehat R_t
&=\operatorname{softmax}\bigl(P_R(\mathbf H_t)\bigr),\\
\widehat F
&=\operatorname{sigmoid}\bigl(P_F(\mathbf H_0)\bigr),\\
\widehat S_t
&=\operatorname{sigmoid}\bigl(P_S(\mathbf H_t)\bigr),\\
\widehat D_t
&=P_D(\mathbf H_t)
\end{aligned}
\label{eq:prediction}
\end{equation}
Here, $P$ denotes a learnable prediction head. The future response prediction uses contextual features at the initial state, which already incorporate the task instruction and spatiotemporal information from the current noisy video.

\subsection{Response Feedback Adapter}
The predicted interaction dynamics are then used to guide object changes during generation. As shown in Fig.~\ref{fig:pipeline}(b) and~(c), the response feedback adapter uses two complementary paths: dynamic conditions guide object feature updates, while predicted displacement guides where to read information across frames.

\paragraph{Constructing interaction conditions.}
We pool initial-state features using role and future response intensity weights:
\begin{equation}
m_j
=
\frac{
\sum_x w_j(x)\mathbf H_0(x)
}{
\sum_x w_j(x)+\varepsilon
}
\label{eq:context}
\end{equation}
Here, $j\in\{r,d,a,F\}$. The role weights are $w_k(x)=\widehat R_0^k(x)$, and the future response weight is $w_F(x)=\max_{k\in\mathcal K}\widehat F^k(x)$, with $\varepsilon$ for numerical stability. This gives the interaction context $M=[m_r,m_d,m_a,m_F]$: the first three vectors summarize the participants, while the fourth emphasizes local regions with stronger future responses.

For each temporal interval, we further convert predicted motion support into a continuous gate $g_t^k(x)\in[0,1]$ and combine it with predicted displacement:
\begin{equation}
p_t^k(x)
=
\left[
g_t^k(x),\;
g_t^k(x)\bar D_t^k(x)
\right]
\label{eq:dynamic}
\end{equation}
Here, $\bar D$ is the clipped displacement prediction. The gate controls the strength of the local dynamics condition, while displacement specifies motion direction and distance. $M$ supplies interaction context, and $p$ carries local motion into the feature update.

\paragraph{Updating object features with a zero-dynamics reference.}
The object branch can produce an output from video features and interaction context even without dynamic input. To make the added object update explicitly depend on predicted dynamics, we hold the features and context fixed and use the zero-dynamics output as a reference:
\begin{equation}
\Delta_t^k
=
f_o(\mathbf H_t,M,p_t^k)
-
f_o(\mathbf H_t,M,0)
\label{eq:objectdelta}
\end{equation}
Here, $k\in\{d,a\}$. Both evaluations share the parameters of $f_o$ and differ only in their dynamic conditions. Thus, $\Delta_t^k$ represents the feature change relative to a common reference after introducing predicted dynamics; it is zero when the local dynamics condition is zero. We then update the video features together with the residual from the robot branch $f_r$:
\begin{equation}
\begin{split}
\mathbf H_t^{\mathrm{rf}}
={}&
\mathbf H_t+f_r(\mathbf H_t,M,p_t^r)\\
&+\eta(\Delta_t^d+\Delta_t^a)
\end{split}
\label{eq:objectdelta-update}
\end{equation}
Here, $\eta>0$ is a learnable scale. The updated $\mathbf H_t^{\mathrm{rf}}$ is then used for cross-frame information aggregation.

\paragraph{Displacement-guided temporal attention.}
After updating features, the model must determine where to read information from the next state to account for object motion. We obtain queries from $\mathbf H_t^{\mathrm{rf}}$ and keys/values from $\mathbf H_{t+1}^{\mathrm{rf}}$, and apply temporal attention within a fixed local window. Predicted displacement specifies a spatial preference within the window, and the motion gate modulates its strength:
\begin{equation}
b_t(x,\delta)
=
\beta
\max_{k\in\{r,o\}}
\left[
g_t^k(x)
\exp\left(
-\frac{\|\delta-\bar D_t^k(x)\|_2^2}{2\tau^2}
\right)
\right]
\label{eq:bias}
\end{equation}
Here, $\delta$ denotes a candidate offset from $x$. At each location, $o$ selects the direct target or affected object with the larger gate, keeping that role's gate and displacement together. $\beta\ge0,\tau>0$ control the bias strength and width, respectively. The bias favors candidate offsets close to the predicted displacement, while the candidate window remains fixed.

For a single attention head, we add the bias to content similarity to obtain cross-frame attention weights:
\begin{equation}
\begin{split}
a_t(x,\delta)
={}&\operatorname{softmax}_{\delta}\Bigl[\\
&\frac{q_t(x)^\top k_{t+1}(x+\delta)}{\sqrt{C_h}}
+b_t(x,\delta)\Bigr]
\end{split}
\label{eq:attention}
\end{equation}
Here, $C_h$ is the head dimension, and softmax normalizes over the fixed candidate window $\mathcal N(x)$. We then aggregate features from the next state using these weights:
\begin{equation}
y_t(x)
=
\sum_{\delta\in\mathcal N(x)}
a_t(x,\delta)\,
v_{t+1}(x+\delta)
\label{eq:aggregation}
\end{equation}
Concatenating the head outputs gives $Z_t$, which updates the features through a residual with controlled magnitude:
\begin{equation}
\widetilde{\mathbf H}_t
=
\mathbf H_t^{\mathrm{rf}}
+
\rho_t\tanh\bigl(W_{\mathrm{up}}Z_t\bigr)
\label{eq:bounded}
\end{equation}
Here, $W_{\mathrm{up}}$ is the output projection, and $\rho_t$ is set according to the current feature magnitude to control the relative strength of the added feedback. The updated features are passed to the remaining backbone layers. As shown in Fig.~\ref{fig:response}(b), content similarity and predicted motion jointly determine the cross-frame attention weights.

\subsection{Training and Generation}
Response supervision grounds structured interaction anticipation in labeled targets, while the video loss optimizes how these predictions are used in generation through the feedback paths. The joint objective is:
\begin{equation}
\mathcal L
=
\mathcal L_{\mathrm{video}}
+
\lambda_R\mathcal L_R
+
\lambda_F\mathcal L_F
+
\lambda_U\mathcal L_U
\label{eq:jointloss}
\end{equation}
Here, $\mathcal L_R$ supervises task-related roles with cross-entropy, $\mathcal L_F$ regresses future response intensity with Smooth L1, and $\mathcal L_U$ combines binary cross-entropy for motion support with Smooth L1 for displacement. Response labels are constructed offline from training videos, and response losses are computed only where valid labels are available.

Inference requires only the initial image and task instruction. At each denoising step, the model predicts interaction responses from the current features and feeds them back into the generation backbone through object feature updates and displacement-guided temporal attention. The response conditions are continually updated as the video is refined.

\section{Experiments}
\label{sec:experiments}
We evaluate ReactWorld on RBench~\cite{ref24_rbench} and EZS-Bench~\cite{ref25_abot} in terms of task completion, interaction plausibility, and visual quality. Component ablations and displacement interventions examine the roles of structured dynamics prediction and its feedback into generation.

\begin{table*}[!t]
	\centering\small
	\caption{RBench category scores. $*$ denotes results from the official Qwen leaderboard.}
	\label{tab:rbench}
	\setlength{\tabcolsep}{4pt}
	\renewcommand{\arraystretch}{1.12}
	\begin{tabular*}{\textwidth}{@{\extracolsep{\fill}}lrrrrrrrrrr@{}}
		\toprule
		\multirow{2}{*}{Method}
		& \multirow{2}{*}{Avg.$\uparrow$}
		& \multicolumn{5}{c}{Task category $\uparrow$}
		& \multicolumn{4}{c}{Robot morphology $\uparrow$}\\
		\cmidrule(lr){3-7}\cmidrule(l){8-11}
		& & Manipulation & Spatial
		& \shortstack{Multi-\\entity}
		& \shortstack{Long-\\horizon}
		& Reasoning
		& \shortstack{Single\\arm}
		& \shortstack{Dual\\arm}
		& Quadruped & Humanoid\\
		\midrule
		Wan2.2-A14B$^*$~\cite{wan2025}
		&69.80&70.90&66.00&92.10&68.10&55.00&68.80&67.80&\textbf{67.00}&\textbf{72.80}\\
		Cosmos3-Nano~\cite{nvidia2026cosmos3}
		&66.33&69.55&78.00&73.40&56.98&63.25&61.65&66.95&60.80&66.35\\
		ABot-PhysWorld~\cite{ref25_abot}
		&57.68&64.40&59.50&67.00&42.93&38.00&61.25&63.60&59.55&62.85\\
		PhysisForcing~\cite{ref03_physisforcing}
		&74.21&75.65&80.00&94.20&74.73&78.13&67.75&67.60&60.85&69.00\\
		\midrule
		Video-FT
		&71.94&73.50&71.50&91.90&70.80&63.80&68.20&68.40&66.90&72.50\\
		\midrule
		ReactWorld
		&\textbf{77.51}&\textbf{80.45}&\textbf{86.00}&\textbf{96.00}&\textbf{78.73}&\textbf{82.13}&\textbf{70.60}&\textbf{69.80}&63.35&70.50\\
		\bottomrule
	\end{tabular*}
\end{table*}

\subsection{Experimental Setup}
\paragraph{Training data and annotations.}
We curate approximately 500K clips from AgiBot~\cite{agibot2025world}, RoVid-X~\cite{ref24_rbench}, and robot videos collected from the Internet. For samples with reliable annotations, we construct interaction dynamics supervision offline, including participant roles, future motion response intensity referenced to the first frame, and motion regions and 2D displacements between adjacent states. We use Qwen3-VL~\cite{bai2025qwen3vl} to assist task-related role assignment, SAM3~\cite{carion2025sam3} to obtain object masks, and CoWTracker~\cite{lai2026cowtracker} to extract point trajectories. Role and motion observations are combined to construct the supervision. Training and validation sets are split by original episode or source video.

\paragraph{Implementation details.}
Videos contain 81 frames at $480\times832$ and 16 FPS. We initialize from Wan2.2-I2V-A14B~\cite{wan2025} and freeze the backbone. We train rank-32 video LoRA, response prediction heads, residual adapters, and the temporal attention branch in BF16 on 8 H100 GPUs. Response prediction and feedback are both applied at the DiT block~12. We use AdamW with a learning rate of $10^{-5}$ and an effective batch size of 32. The first 64 steps warm up LoRA and the prediction heads, followed by joint training with the feedback modules. The output projection $W_{\mathrm{up}}$ of temporal attention is initialized to zero. Further implementation details are provided in the supplementary material.

\paragraph{Evaluation setup.}
RBench~\cite{ref24_rbench} covers five task categories and four robot morphologies. We report all nine category scores and their average, and analyze robot and object stability (RSS), physical--semantic plausibility (PSS), and task-adherence consistency (TAC) on the 400 examples spanning the four morphologies. EZS-Bench~\cite{ref25_abot} reports embodied domain performance (Domain), perceptual quality (Quality), and their average. RBench results are evaluated with Qwen3-VL-235B-A22B-Instruct~\cite{bai2025qwen3vl}. For EZS-Bench, Domain is evaluated with Qwen2.5-VL-72B-Instruct~\cite{bai2025qwen25vl}, while Quality is computed using the benchmark's eight video quality metrics.

\paragraph{Baselines.}
We compare with Wan2.2-A14B~\cite{wan2025}, Cosmos3-Nano~\cite{nvidia2026cosmos3}, ABot-PhysWorld~\cite{ref25_abot}, and PhysisForcing~\cite{ref03_physisforcing}. We also include Video-FT, a standard video fine-tuning baseline using the same initialization, data, and training budget as ReactWorld.

\begin{figure*}[!t]
\centering
\includegraphics[width=\textwidth]{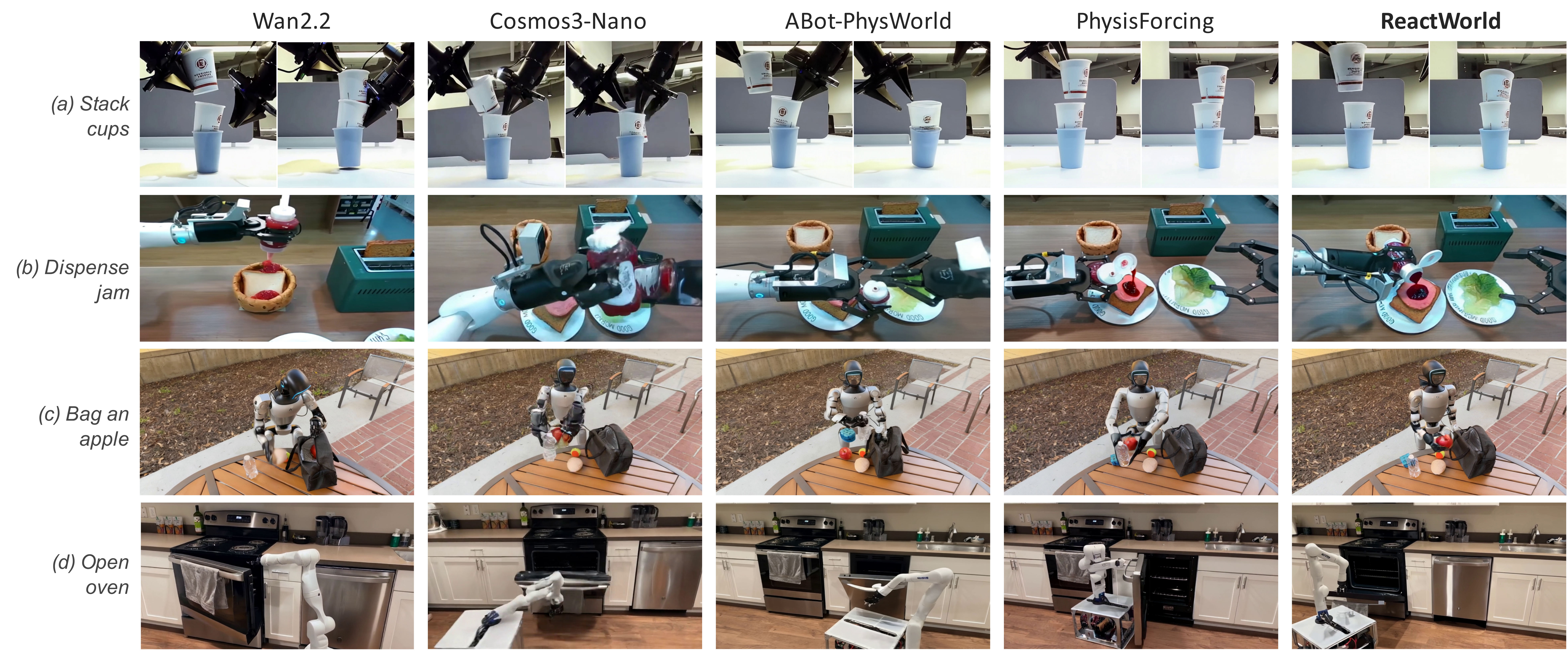}
\caption{\textbf{Embodied task generation across methods.} Columns show Wan2.2~\cite{wan2025}, Cosmos3-Nano~\cite{nvidia2026cosmos3}, ABot-PhysWorld~\cite{ref25_abot}, PhysisForcing~\cite{ref03_physisforcing}, and ReactWorld. Rows show stacking cups, dispensing jam, bagging apples, and opening an oven.}
\label{fig:qualitative}
\end{figure*}

\subsection{Embodied Video Generation Results}
\paragraph{RBench evaluation.}
In Tab.~\ref{tab:rbench}, ReactWorld achieves an average score of 77.51, outperforming Video-FT (71.94) and improving over PhysisForcing by 3.29 points. It leads all five task categories, with gains over PhysisForcing of 6.00, 4.80, 4.00, and 4.00 points in spatial relations, manipulation, long-horizon planning, and visual reasoning, respectively, indicating better adherence to spatial constraints and task requirements.

ReactWorld outperforms PhysisForcing in all nine categories and Video-FT in seven. Its gains over Video-FT reach 14.50 points in spatial relations and 18.33 points in visual reasoning, supporting the effectiveness of structured interaction modeling for spatial reasoning and task understanding.

\paragraph{Interaction quality.}
Compared with PhysisForcing, ReactWorld improves RSS from 11.24 to 11.51 and PSS from 4.52 to 4.62. Both metrics improve across all four morphologies, indicating gains in participant stability and the plausibility of object responses. Averaged across the four morphologies, task completion (TC) increases from 83.58 to 85.18, visual quality (VQ) from 49.03 to 51.95, and task-adherence consistency (TAC) from 4.17 to 4.20.

\begin{table}[!t]
	\centering\small
	\caption{EZS-Bench196 results. }
	\label{tab:ezs}
	\setlength{\tabcolsep}{4pt}
	\renewcommand{\arraystretch}{1.10}
	\begin{tabular*}{\columnwidth}{@{\extracolsep{\fill}}lrrr@{}}
		\toprule
		Method & Domain$\uparrow$ & Quality$\uparrow$ & Avg.$\uparrow$\\
		\midrule
		Wan2.2-A14B~\cite{wan2025}
		&77.42&76.89&77.16\\
		Cosmos3-Nano~\cite{nvidia2026cosmos3}
		&84.32&75.60&79.96\\
		ABot-PhysWorld~\cite{ref25_abot}
		&83.66&\textbf{76.94}&80.30\\
		PhysisForcing~\cite{ref03_physisforcing}
		&84.49&76.58&80.54\\
		\midrule
		Video-FT
		&83.45&76.64&80.05\\
		\midrule
		ReactWorld
		&\textbf{86.98}&76.75&\textbf{81.87}\\
		\bottomrule
	\end{tabular*}
\end{table}

\paragraph{EZS-Bench evaluation.}
ReactWorld achieves a Domain score of 86.98, exceeding Video-FT and PhysisForcing by 3.53 and 2.49 points, respectively, and an average score of 81.87 (Tab.~\ref{tab:ezs}). Its Quality score of 76.75 is also slightly higher than both baselines, indicating improved embodied domain performance while maintaining perceptual quality.

\begin{table}[!tb]
\centering\small
\caption{Human evaluation. We report mean ratings on a five-point scale. Parentheses give Holm-adjusted $p$-values from paired $t$-tests against ReactWorld.}
\label{tab:human}
\setlength{\tabcolsep}{2pt}\renewcommand{\arraystretch}{1.12}
\begin{tabular*}{\columnwidth}{@{\extracolsep{\fill}}lccc@{}}
\toprule
Method & \shortstack{Task\\progress$\uparrow$} & \shortstack{Contact\\plausibility$\uparrow$} & \shortstack{Visual\\quality$\uparrow$}\\
\midrule
Video-FT &3.74 ($p{=}0.063$)&3.24 ($p{<}0.001$)&3.92 ($p{=}0.211$)\\
PhysisForcing &3.70 ($p{=}0.006$)&3.78 ($p{=}0.140$)&\textbf{4.16} ($p{=}0.749$)\\
ReactWorld &\textbf{4.08}&\textbf{4.06}&4.12\\
\bottomrule
\end{tabular*}
\end{table}

\paragraph{Human evaluation.}
We invite 50 evaluators to rate task progress, contact plausibility, and visual quality on a five-point scale. As shown in Tab.~\ref{tab:human}, ReactWorld achieves the highest scores for task progress and contact plausibility, at 4.08 and 4.06, respectively. Its contact plausibility is significantly higher than Video-FT ($p<0.001$), while its visual quality score of 4.12 indicates that it combines high visual quality with more plausible interactions.

\paragraph{Qualitative results.}
\begingroup\widowpenalty=10000
Fig.~\ref{fig:qualitative} presents results for stacking cups, dispensing jam, bagging apples, and opening an oven. For jam dispensing, some baselines do not clearly establish the relationship between the container opening, the source of the jam, and its destination. ReactWorld more clearly depicts the correspondence between the bottle opening and the receiving target. Its oven-opening result shows the open door and interior structure, while its cup-stacking result is more physically plausible. These differences in action--response correspondence and task states are consistent with the gains in task completion and physical plausibility.\par
\endgroup

\begin{figure}[!t]
	\centering
	\includegraphics[width=\columnwidth]{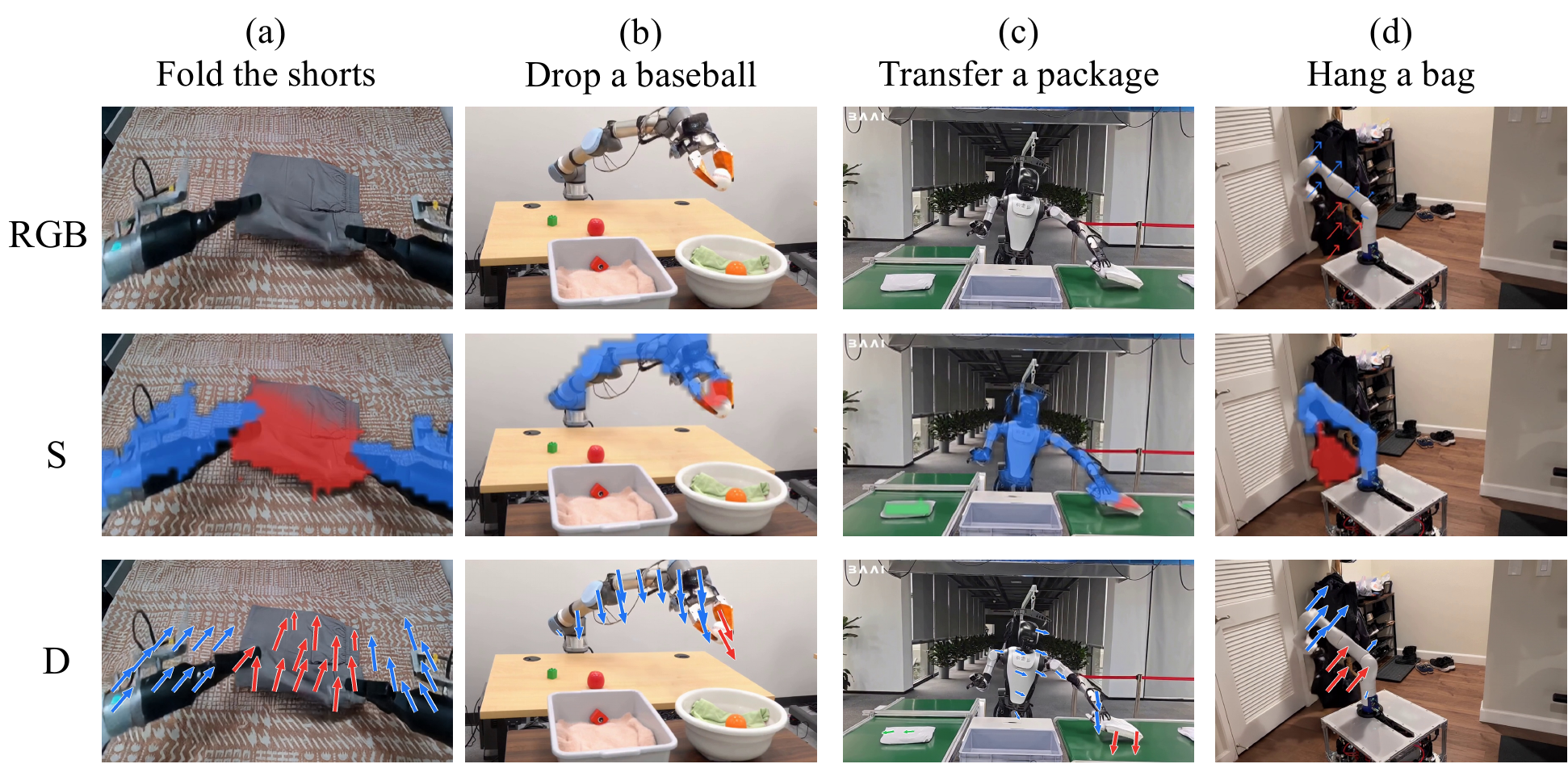}
	\caption{\textbf{Interaction dynamics across tasks.}
		Columns show (a) folding shorts,
		(b) dropping a baseball into a basin,
		(c) transferring a package, and
		(d) hanging a bag.
		Rows show generated frames, predicted motion support $S$,
		and displacement $D$.}
	\label{fig:dynamics}
\end{figure}

\paragraph{Predicted interaction dynamics.}
Fig.~\ref{fig:dynamics} visualizes motion support and displacement
predicted from intermediate backbone features across four tasks.
The support maps indicate where local motion is anticipated,
while displacement vectors describe its direction and magnitude.
Together, they provide an interpretable view of the local
interaction dynamics that ReactWorld uses to guide generation.

\subsection{Ablations and Feedback Analysis}
\paragraph{Structured supervision and generative feedback.}
Tab.~\ref{tab:ablation} compares response supervision alone with feedback from predicted dynamics. Heads-only adds only response supervision, and Heads+Adapter adds dynamics-conditioned feature updates. Full w/o bias further introduces content-based temporal aggregation, while Full additionally uses displacement bias. All variants use the same initialization, data, and training budget; all except Video-FT use the same response supervision.

Full improves PSS and Domain by 0.49 and 2.87 points over Heads-only, supporting the benefit of dynamics feedback. It also exceeds Full w/o bias by 0.25 and 1.60 points, supporting displacement-based guidance beyond response updates and content attention.

\begin{table}[!t]
\centering\small
\caption{Ablation of response supervision and generative feedback.}
\label{tab:ablation}
\setlength{\tabcolsep}{2.5pt}\renewcommand{\arraystretch}{1.10}
\begin{tabular*}{\columnwidth}{@{\extracolsep{\fill}}lrrrr@{}}
\toprule
\multirow{2}{*}{Variant} & \multicolumn{2}{c}{RBench} & \multicolumn{2}{c}{EZS-Bench}\\
\cmidrule(lr){2-3}\cmidrule(l){4-5}
 & PSS$\uparrow$ & TAC$\uparrow$ & Domain$\uparrow$ & Quality$\uparrow$\\
\midrule
Video-FT &4.01&3.81&83.45&76.64\\
Heads-only &4.13&3.90&84.11&76.66\\
Heads+Adapter &4.24&4.12&84.73&76.69\\
Full w/o bias &4.37&4.15&85.38&76.71\\
\midrule
Full (ReactWorld) &\textbf{4.62}&\textbf{4.20}&\textbf{86.98}&\textbf{76.75}\\
\bottomrule
\end{tabular*}
\end{table}

\paragraph{Local responses and the zero-dynamics reference.}
Tab.~\ref{tab:reference} shows that pooling context by future response intensity, rather than over whole role regions, improves PSS and Domain by 0.31 and 2.13 points, respectively. This indicates that emphasizing regions with stronger responses helps generate more plausible interactions. Without the zero-dynamics reference, the two scores remain 0.11 and 1.03 points below the full model even after matching the object update magnitude, supporting the benefit of the reference at comparable update magnitudes.

\begin{table}[!t]
\centering\small
\caption{Ablation of local responses and the zero-dynamics reference.}
\label{tab:reference}
\setlength{\tabcolsep}{2.5pt}\renewcommand{\arraystretch}{1.10}
\begin{tabular*}{\columnwidth}{@{\extracolsep{\fill}}lrrrr@{}}
\toprule
\multirow{2}{*}{Variant} & \multicolumn{2}{c}{RBench} & \multicolumn{2}{c}{EZS-Bench}\\
\cmidrule(lr){2-3}\cmidrule(l){4-5}
 & PSS$\uparrow$ & TAC$\uparrow$ & Domain$\uparrow$ & Quality$\uparrow$\\
\midrule
Full &\textbf{4.62}&\textbf{4.20}&\textbf{86.98}&\textbf{76.75}\\
\shortstack[l]{Whole-entity\\context} &4.31&4.14&84.85&76.71\\
No zero reference &4.43&4.12&85.12&76.68\\
Amplitude-matched &4.51&4.16&85.95&76.72\\
\bottomrule
\end{tabular*}
\end{table}

\begin{figure}[!t]
\centering
\includegraphics[width=\columnwidth]{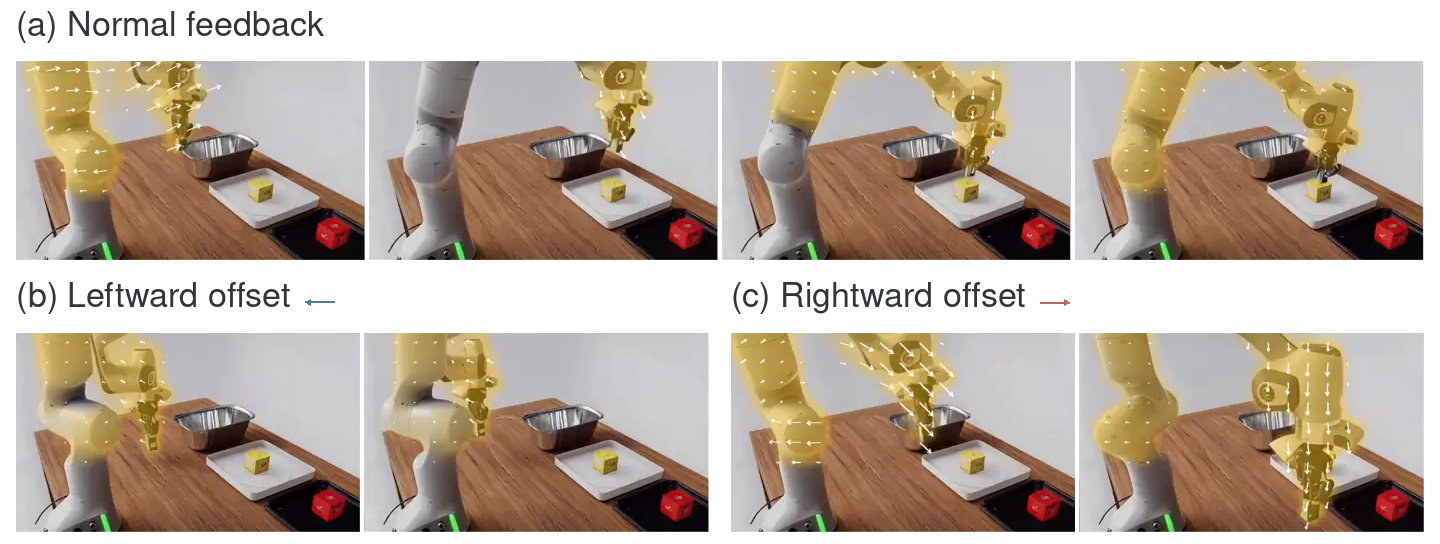}
\caption{\textbf{Directional interventions on feedback displacement.} (a) Normal feedback. (b,c) Leftward and rightward offsets applied to the feedback displacement input shift the robot arm's motion in the corresponding directions.}
\label{fig:motion}
\end{figure}

\paragraph{Directional interventions on feedback displacement.}
The preceding ablations assess the contribution of the feedback design to generation quality. We further examine the correspondence between feedback displacement and generated motion. With short task descriptions, we apply leftward or rightward offsets to the displacement conditions entering the feedback module. As shown in Fig.~\ref{fig:motion}, the robot arm's motion and approach position shift in the corresponding directions under different displacement conditions. These examples show a qualitative correspondence between feedback displacement and generated motion, suggesting the potential to guide local interactions through internal feedback conditions.

\begin{figure}[!t]
\centering
\includegraphics[width=\columnwidth]{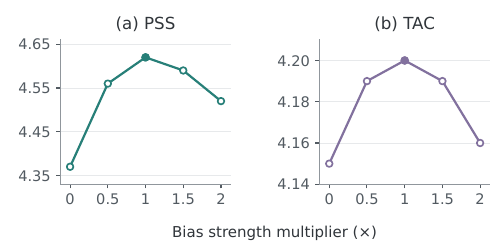}
\caption{\textbf{Displacement bias strength.} Filled markers denote the default setting. A multiplier of zero disables positional bias while retaining object feature updates and content-based temporal attention.}
\label{fig:beta}
\end{figure}

\paragraph{Displacement bias strength.}
With the full model fixed, we scale displacement bias strength by 0, 0.5, 1, 1.5, and 2 (Fig.~\ref{fig:beta}). The default strength improves PSS and TAC by 0.25 and 0.05 points over disabling the bias. Scores remain close from 0.5 to 1.5 times the default and decrease at 2 times, supporting moderate positional guidance for embodied video generation.

%

\section{Conclusion}
We presented ReactWorld, which anticipates participant roles, future motion response intensity, and interval motion in the generation backbone's feature space, and feeds the predicted dynamics back into video generation. Differences relative to a zero-dynamics reference guide object feature updates, while displacement biases guide cross-frame information retrieval. Experiments on RBench and EZS-Bench show that this design improves task completion and the physical plausibility of interactions.

\paragraph{Limitations and future work.}
ReactWorld currently anticipates interactions using a 2D response representation, leaving room for explicit modeling of 3D interaction relations. We will explore temporally consistent 3D motion representations and rehearsal to improve spatial consistency.
	
\section*{LLM Usage Statement}
ChatGPT (OpenAI) assisted manuscript drafting, English editing, literature organization, figure and LaTeX preparation. Qwen3-VL assisted task-related role annotation. Qwen3-VL-235B-A22B-Instruct served as the RBench evaluator, and Qwen2.5-VL-72B-Instruct was used for Domain evaluation on EZS-Bench, as described in the experimental setup.
	
	{\small
		\bibliographystyle{ieeenat_fullname}
		\bibliography{references}
	}
	
\end{document}


\renewcommand{\thesection}{S\arabic{section}}
\renewcommand{\theequation}{S\arabic{equation}}
\renewcommand{\thetable}{S\arabic{table}}
\renewcommand{\thefigure}{S\arabic{figure}}
\twocolumn[{%
  {\centering\Large
  \textbf{\thetitle}\\
  \vspace{0.5em}Supplementary Material\par}
  \vspace{1.0em}
  \noindent\begin{minipage}{\linewidth}
  \centering
  \includegraphics[width=0.95\linewidth]{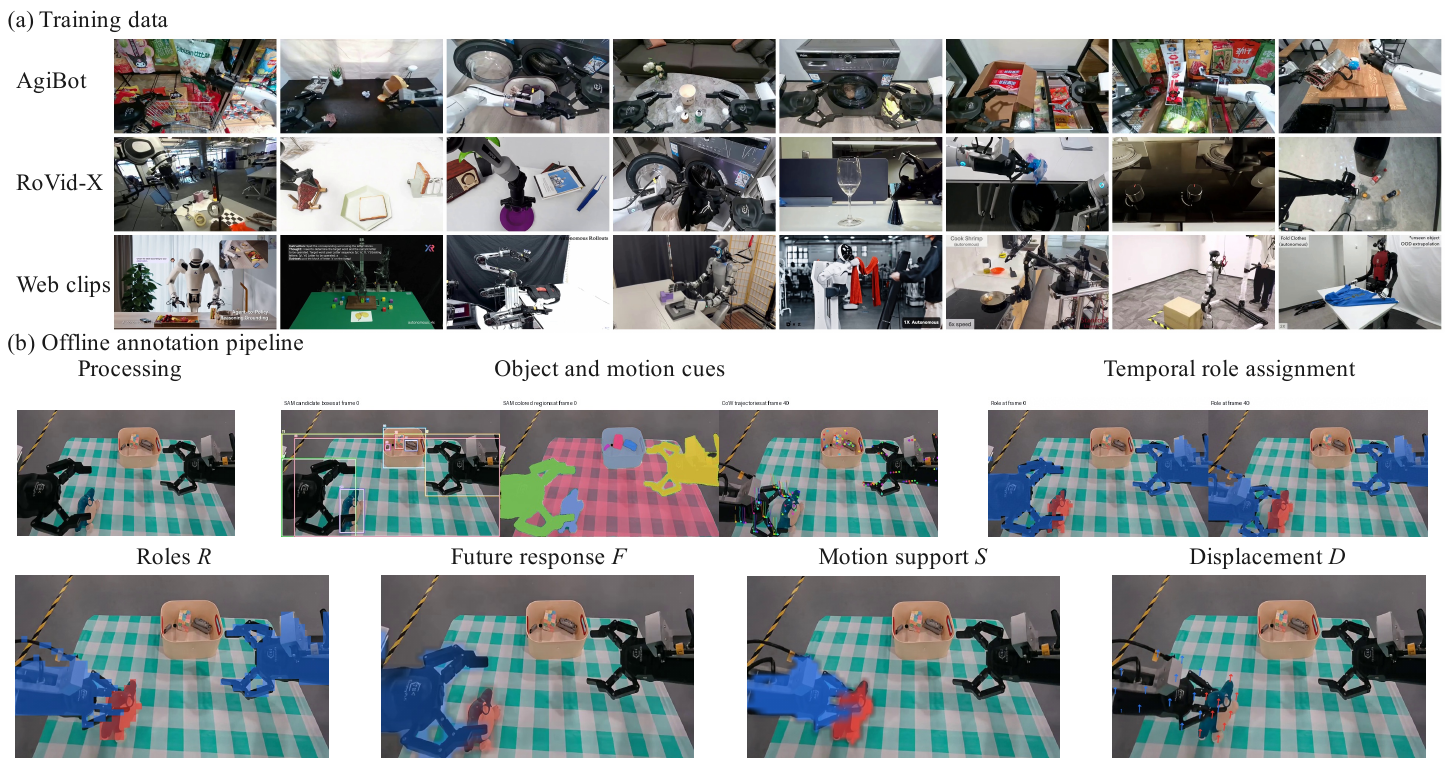}\par
  \FigureCaption{\textbf{Training data and offline interaction supervision.}
    (a) Representative initial frames from AgiBot, RoVid-X, and web robot videos.
    (b) Clip preparation, mask and trajectory cues, temporal role assignment, and the resulting $R$, $F$, $S$, and $D$ targets.}{fig:annotation}
\end{minipage}\par

  \vspace{6pt}
}]
\begingroup
\frenchspacing
\lefthyphenmin=2\relax
\righthyphenmin=3\relax
\section{Training Data and Interaction Annotation}
\label{sec:data}
Our training collection contains approximately 500K clips from AgiBot~\cite{agibot2025world}, RoVid-X~\cite{ref24_rbench}, and web robot videos (Fig.~\ref{fig:annotation}(a)). We prepare the public-dataset clips and web clips separately, then construct shared interaction supervision. Table~\ref{tab:data} summarizes the training videos by source and supervision.
Figure~\ref{fig:training-distribution} shows the training task distribution across all three sources. Pick-and-place accounts for the largest share. We give long-tail tasks higher sampling weights during training.
\suppressfloats[t]
\begin{figure}[t]
  \centering
  \includegraphics[width=\linewidth,height=\linewidth,keepaspectratio]{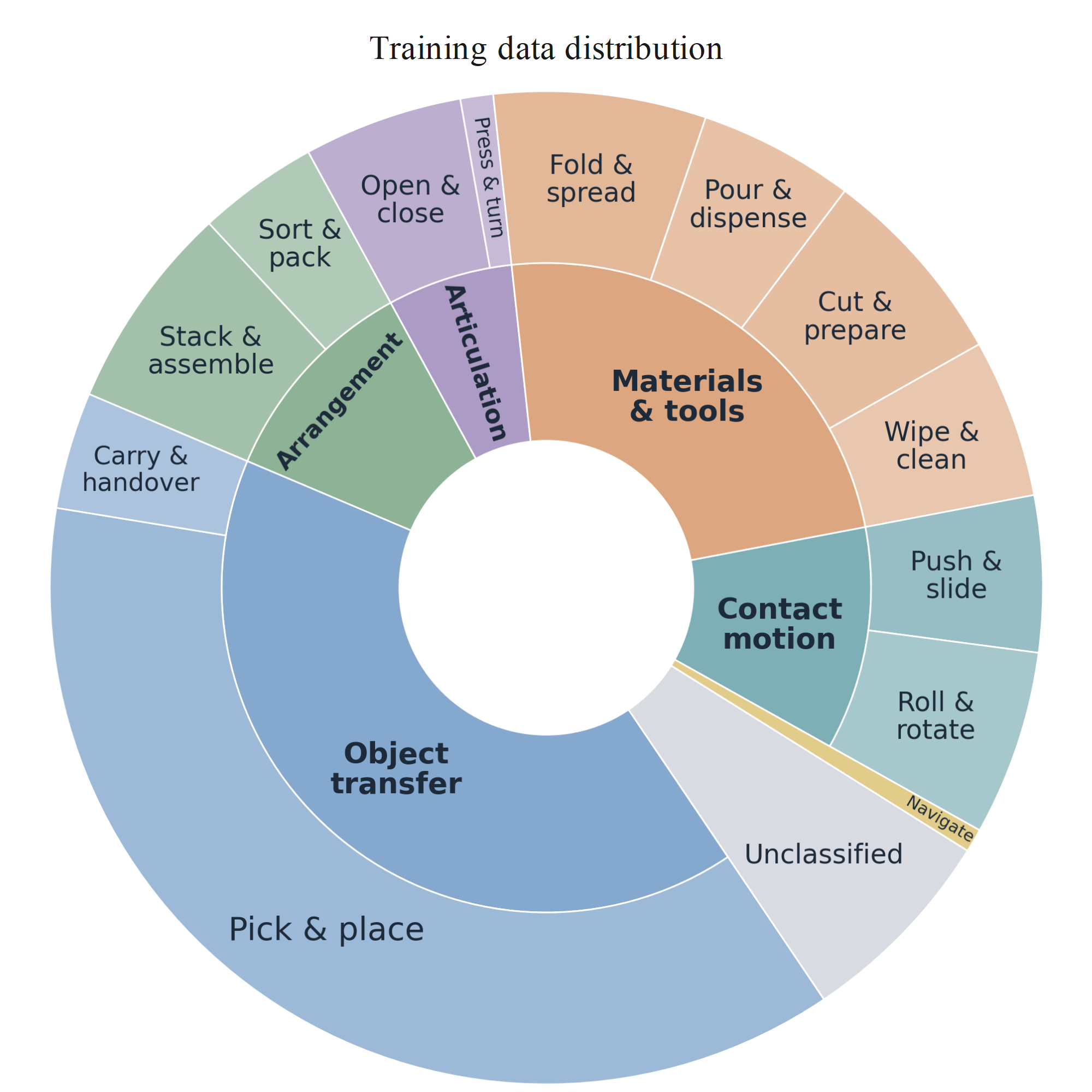}
  \captionsetup{skip=4pt}
  \caption{\textbf{Training data distribution.}
    The inner ring shows task groups, and the outer ring shows individual tasks.}
  \label{fig:training-distribution}
\end{figure}

\subsection{AgiBot and RoVid-X}
We retain the correspondence between episodes, source videos, and task descriptions. RoVid-X is filtered by action--object combinations to reduce repeated descriptions and improve coverage of sparse tasks. The original descriptions provide the training text. After checking decoding, duration, and resolution, accepted clips of \PublicClipRange{} seconds are normalized by uniformly sampling 81 frames over the full accepted clip and applying a common spatial transform to obtain \mbox{$480\times832$} videos. We extract the first frame from the normalized video and retain source-frame indices and crop parameters for annotation alignment.
\par\endgroup

\makeatletter
\IfFileExists{supplement/figures/figS1_annotation.pdf}{%
  \if@firstcolumn
    \newpage
    \if@firstcolumn\else
    \null\par
    \fi
  \fi
}{}
\makeatother
\clearpage
\begingroup
\setlength{\textfloatsep}{6pt plus 1pt minus 1pt}

\begin{table}[!t]
\begingroup
\centering
\captionsetup{type=table,position=top,skip=2pt}
\caption{Training videos by source and supervision.}\label{tab:data}
\small\setlength{\tabcolsep}{2pt}\renewcommand{\arraystretch}{1.07}
\begin{tabular*}{\columnwidth}{@{\extracolsep{\fill}}lrrr@{}}
\toprule Source & Videos & Joint RFU & Video-only\\\midrule
AgiBot & 32,492 & 22,492 & 10,000\\
RoVid-X & 243,869 & 193,878 & 49,991\\
Web & 223,771 & 218,186 & 5,585\\\midrule
\textbf{Total} & \textbf{500,132} & \textbf{434,556} & \textbf{65,576}\\\bottomrule
\end{tabular*}
\par\endgroup

\end{table}
\RWRightTable{
\begingroup
\centering
\captionsetup{type=table,position=top,skip=3pt}
\caption{Model and training settings.}
\label{tab:config}
\small\setlength{\tabcolsep}{3pt}\renewcommand{\arraystretch}{1.03}
\begin{tabular*}{\columnwidth}{@{\extracolsep{\fill}}p{.46\columnwidth}p{.49\columnwidth}@{}}
\toprule Setting & Value\\\midrule
Backbone & Wan2.2-I2V-A14B (frozen)\\
Video format & 81 frames, $480\times832$, \OutputFPS~FPS\\
DiT insertion block & \InterfaceBlock{}\\
Local attention window & $7\times7$\\
DiT grid / hidden width & $21\times30\times52$ / 5120\\
Video LoRA rank & 32\\
Optimizer / learning rate & AdamW, $10^{-5}$\\
Effective batch size & 32\\
LoRA and head warm-up & 64 steps\\
Hardware / precision & $8\times$H100, BF16\\\bottomrule
\end{tabular*}
\par\endgroup
}
\begingroup
\frenchspacing
\lefthyphenmin=2\relax
\righthyphenmin=3\relax
\subsection{Web Clip Selection and Annotation}
Web material is collected from robot manufacturers, public project pages, and video sources, with provenance and source identity retained. The recorded automatic proposal configuration scans windows of \WebWindow{} seconds with \WebOverlap{} seconds of overlap and refines continuous-action candidates of \WebClipRange{} seconds. Manual selection follows a complete interaction or coherent subtask, with approximately ten seconds as a reference. Clips retain the pre-contact state, key manipulation, and visible outcome. Shot transitions define boundaries, and a fixed crop preserves the robot, manipulated objects, and necessary motion space.

\mbox{Qwen3.8-27B}~\cite{qwen38_27b} examines timestamped frames spanning each candidate clip to propose action descriptions and flag uncertainties. Human reviewers confirm, revise, or reject these proposals. Descriptions refer only to visible actions after the first frame. Reviewers label a grasp when it is visible and mark a placement as complete after release. Camera and robot motion are distinguished, and failed or uncertain outcomes are retained. Clip boundaries and English descriptions are checked together.

Selected web clips vary in duration before normalization. As with the public datasets, we uniformly sample 81 frames over the full clip and normalize the spatial resolution to \mbox{$480\times832$}. Samples without reliable response labels contribute to video training with the corresponding response losses disabled. Splits are grouped by source video or episode. Text and spatial variants retain their source clip identity.

\subsection{Shared Supervision and Quality Control}
Task parsing and global object proposals provide localization cues (Fig.~\ref{fig:annotation}(b)). Multiplex masks from SAM3~\cite{carion2025sam3} are fused and deduplicated into object candidates. CoWTracker~\cite{lai2026cowtracker} trajectories and visibility, together with object motion and robot association, support role assignment. The four roles are background, robot, direct target, and affected object.

$F$ associates subsequent local responses with the first frame. The $30\times52$ grid has 21 states at frames $0,4,\ldots,80$ and 20 intervals. $S$ and $D$ use interval-start coordinates and represent motion support and grid-unit displacement. \MotionMapping{} $S$ distinguishes moving, known static, and ignored cells. Reliability $Q$ controls validity or loss weights. Static support requires a valid observation. Numerical values, endpoints, spatial alignment, and sampled overlays are checked before export.
\par\endgroup

\section{Response Feedback and Implementation Details}
\label{sec:supp-feedback}
The response predictor and feedback modules operate at DiT block~\InterfaceBlock{}. Table~\ref{tab:config} summarizes the model and training settings.

\paragraph{Interaction context and local dynamics.}
At each denoising step, we pool the initial-state features into three role vectors and one future-response vector. We share the context $M=[m_r,m_d,m_a,m_F]$ across time. For each interaction role $k\in\{r,d,a\}$, we construct the local condition as
\begin{equation}
\begin{aligned}
g_t^k(x)&=\operatorname{clip}\!\left(\frac{\widehat S_t^k(x)-s_0}{1-s_0},0,1\right),\\
\bar D_t^k(x)&=\operatorname{clip}\!\left(\widehat D_t^k(x),-d_{\max},d_{\max}\right),\\
p_t^k(x)&=[g_t^k(x),\,g_t^k(x)\bar D_t^k(x)].
\end{aligned}\label{eq:supp-dynamic}
\end{equation}
Here $s_0$ is the support threshold, and $d_{\max}$ bounds each displacement component in grid-cell units. We set $p_{T-1}^k=0$ at the final state.

\paragraph{Adapters and the zero-dynamics reference.}
The robot and object adapters $f_r$ and $f_o$ have the form
\begin{equation}
\begin{aligned}
\xi_\ell&=W_\ell^\downarrow\operatorname{LN}(H)+\psi_\ell(M)+W_\ell^p p,\\
f_\ell(H,M,p)&=W_\ell^\uparrow\operatorname{SiLU}(\xi_\ell),\quad\ell\in\{r,o\}.
\end{aligned}\label{eq:supp-adapter}
\end{equation}
The map $\psi_\ell$ projects the context into the bottleneck and broadcasts it across the grid. The direct target and affected object share $f_o$. The object increment is $f_o(H,M,p)-f_o(H,M,0)$ with the same features and context in both terms. It vanishes at $p=0$. A nonzero gate can produce an update even at zero displacement.

\twocolumn[{%
  {\centering
\captionsetup{type=table,position=top,skip=3pt}\caption{Detailed RBench results by robot morphology.}\label{tab:morphology}
\small\setlength{\tabcolsep}{3pt}\renewcommand{\arraystretch}{1.08}
\begin{tabular*}{\textwidth}{@{\extracolsep{\fill}}llrrrrrrrr@{}}
\toprule
Embodiment & Method & RSS & PSS & TAC & MS & TC & VQ & \shortstack{Robot\\motion} & \shortstack{Target\\motion}\\\midrule
\multirow{2}{*}{Single arm} & PhysisForcing & 11.22 & 4.47 & 4.43 & \textbf{0.9574} & 86.20 & 49.30 & 0.3903 & 0.1714\\
 & ReactWorld & \textbf{11.59} & \textbf{4.59} & \textbf{4.50} & 0.9537 & \textbf{88.60} & \textbf{52.60} & 0.3777 & 0.1641\\
\addlinespace[2pt]
\multirow{2}{*}{Dual arm} & PhysisForcing & 11.37 & 4.53 & \textbf{4.36} & 0.9106 & 86.10 & 49.10 & 0.2273 & 0.1733\\
 & ReactWorld & \textbf{11.70} & \textbf{4.64} & 4.34 & \textbf{0.9132} & \textbf{87.20} & \textbf{52.40} & 0.2184 & 0.1568\\
\addlinespace[2pt]
\multirow{2}{*}{Quadruped} & PhysisForcing & 10.43 & 4.39 & 3.76 & 0.8340 & 76.90 & 44.80 & 0.3632 & 0.2124\\
 & ReactWorld & \textbf{10.66} & \textbf{4.51} & \textbf{3.84} & \textbf{0.8409} & \textbf{79.40} & \textbf{47.30} & 0.3454 & 0.1775\\
\addlinespace[2pt]
\multirow{2}{*}{Humanoid} & PhysisForcing & 11.92 & 4.68 & \textbf{4.13} & 0.9254 & 85.10 & 52.90 & 0.1548 & 0.1380\\
 & ReactWorld & \textbf{12.10} & \textbf{4.72} & 4.12 & \textbf{0.9300} & \textbf{85.50} & \textbf{55.50} & 0.1552 & 0.1319\\
\bottomrule\end{tabular*}
\par}
  \vspace{8pt}
}]
\RWRightTable{
\begingroup
\centering
\captionsetup{type=table,position=top,skip=3pt}\caption{Mean interaction scores across four morphologies.}\label{tab:shape-average}
\small\setlength{\tabcolsep}{4pt}\renewcommand{\arraystretch}{1.05}
\begin{tabular*}{\columnwidth}{@{\extracolsep{\fill}}lrrr@{}}
\toprule Method &RSS&PSS&TAC\\\midrule
Cosmos3-Nano &11.08&4.44&4.01\\
ABot-PhysWorld &11.29&4.25&3.81\\
PhysisForcing &11.24&4.52&4.17\\
ReactWorld &\textbf{11.51}&\textbf{4.62}&\textbf{4.20}\\\bottomrule
\end{tabular*}
\par\endgroup
}
\paragraph{Object selection and fixed-window attention.}
At each location, we take both the gate and displacement from the object role with the larger gate.
\begin{equation}
\begin{aligned}
k^\star&=\operatorname*{arg\,max}_{k\in\{d,a\}}g_t^k(x),\\
(g_t^o,\bar D_t^o)&=(g_t^{k^\star},\bar D_t^{k^\star}).
\end{aligned}\label{eq:supp-object-selection}
\end{equation}
Queries come from the updated features at state $t$, and keys and values come from the next state. Within the fixed window, we add the larger of the gate-weighted robot and object Gaussian terms to the content logits before softmax. Zero gates recover content-only attention.

\paragraph{Bounded cross-frame feedback.}
Let $Z_t$ concatenate the attention-head outputs. We scale the residual by the current feature magnitude at each location.
\begin{equation}
\begin{aligned}
\rho_t(x)&=\kappa\,\operatorname{sg}\!\left[\operatorname{RMS}_C(\mathbf H_t^{\mathrm{rf}}(x))\right],\\
\widetilde{\mathbf H}_t(x)&=\mathbf H_t^{\mathrm{rf}}(x)+\rho_t(x)\tanh\!\left(W_{\mathrm{up}}Z_t(x)\right).
\end{aligned}\label{eq:supp-bounded}
\end{equation}
We compute the RMS over channels and apply stop-gradient to this magnitude reference. The output projection $W_{\mathrm{up}}$ is initialized to zero. The final state skips cross-frame aggregation.

\paragraph{Response supervision and training.}
We use cross-entropy for roles, binary cross-entropy for motion support, and Smooth L1 for future response and displacement. Each response loss is computed only where the corresponding annotations are valid. Video-only clips contribute to the video objective. After warming up the video LoRA and response heads, we train them jointly with the response adapters and temporal attention branch. At inference, the model takes the initial frame and task instruction.

\section{Evaluation Protocols and Detailed Results}
\subsection{RBench Results by Morphology}
RBench~\cite{ref24_rbench} covers five task categories and four robot morphologies. We use Qwen3-VL-235B-A22B-Instruct~\cite{bai2025qwen3vl} for RBench evaluation. The interaction analysis uses 400 examples spanning the four morphologies.

Table~\ref{tab:morphology} reports robot and object stability (RSS), physical--semantic plausibility (PSS), task-adherence consistency (TAC), motion smoothness (MS), task completion (TC), and visual quality (VQ). Robot and target motion magnitudes are descriptive statistics. ReactWorld scores higher than PhysisForcing on RSS, PSS, TC, and VQ for each morphology.

Table~\ref{tab:shape-average} reports the unweighted mean across morphologies for RSS, PSS, and TAC.

\subsection{EZS-Bench196 Results}
For EZS-Bench196~\cite{ref25_abot}, Qwen2.5-VL-72B-Instruct~\cite{bai2025qwen25vl} evaluates Domain, and the benchmark's eight video-quality metrics determine Quality. Avg.\ is the mean of Domain and Quality. Table~\ref{tab:ezs-quality} reports the eight components on a 0--1 scale and the aggregate scores on a 0--100 scale.

\section{Human Evaluation}
\label{sec:human-detail}
Fifty evaluators rated Video-FT, PhysisForcing, and ReactWorld on task progress, contact plausibility, and visual quality using a five-point scale. Each evaluator contributed one score per method and criterion, yielding 450 ratings.

Table~\ref{tab:human-detail} reports means and sample standard deviations across evaluators. We pair ratings by evaluator for two-sided $t$-tests comparing ReactWorld with each baseline on each criterion ($n=50$, 49 degrees of freedom). We apply the Holm correction across all six comparisons.

At the adjusted $0.05$ threshold, ReactWorld scores higher than Video-FT on contact plausibility ($p<0.001$) and higher than PhysisForcing on task progress ($p=0.006$).

\twocolumn[{%
  \captionsetup{skip=4pt}
\begingroup\centering
\captionsetup{type=table}
\caption{Human evaluation on a five-point scale. Values are mean $\pm$ sample SD. The $p$-values come from two-sided paired $t$-tests against ReactWorld and are Holm-adjusted across six comparisons.}
\label{tab:human-detail}
\small
\setlength{\tabcolsep}{5pt}
\renewcommand{\arraystretch}{1.13}
\begin{tabular*}{\textwidth}{@{\extracolsep{\fill}}lcccccc@{}}
\toprule
\multirow{2}{*}{Method} & \multicolumn{2}{c}{Task progress$\uparrow$} & \multicolumn{2}{c}{Contact plausibility$\uparrow$} & \multicolumn{2}{c}{Visual quality$\uparrow$}\\
\cmidrule(lr){2-3}\cmidrule(lr){4-5}\cmidrule(l){6-7}
& Mean$\pm$SD & $p$ & Mean$\pm$SD & $p$ & Mean$\pm$SD & $p$\\
\midrule
Video-FT & $3.74\pm0.56$ & $0.063$ & $3.24\pm0.72$ & $<0.001$ & $3.92\pm0.53$ & $0.211$\\
PhysisForcing & $3.70\pm0.61$ & $0.006$ & $3.78\pm0.62$ & $0.140$ & $\mathbf{4.16}\pm0.55$ & $0.749$\\
ReactWorld & $\mathbf{4.08}\pm0.70$ & -- & $\mathbf{4.06}\pm0.62$ & -- & $4.12\pm0.69$ & --\\
\bottomrule
\end{tabular*}
\par\endgroup

  \vspace{9pt}
}]
\begin{table}[!t]
  \begingroup
\centering
\captionsetup{type=table,position=top,skip=3pt}\caption{EZS-Bench quality components and overall scores.}\label{tab:ezs-quality}
\small\setlength{\tabcolsep}{3pt}\renewcommand{\arraystretch}{1.05}
\begin{tabular*}{\columnwidth}{@{\extracolsep{\fill}}lrr@{}}
\toprule Metric &Video-FT&ReactWorld\\\midrule
Aesthetic quality &\VftAesthetic&\OursAesthetic\\
Imaging quality &\VftImaging&\OursImaging\\
Motion smoothness &\VftSmoothness&\OursSmoothness\\
Background consistency &\VftBackground&\OursBackground\\
Subject consistency &\VftSubject&\OursSubject\\
Overall consistency &\VftOverall&\OursOverall\\
I2V background &\VftIbg&\OursIbg\\
I2V subject &\VftIsub&\OursIsub\\\midrule
Quality &76.64&\textbf{76.75}\\
Domain &83.45&\textbf{86.98}\\
Avg. &80.05&\textbf{81.87}\\\bottomrule
\end{tabular*}\par\endgroup

\end{table}
\RWRightTable{%
\begingroup
\centering
\captionsetup{type=table,position=top,skip=2pt}\caption{Component configurations.}\label{tab:ablation-config}
\small\setlength{\tabcolsep}{2pt}\renewcommand{\arraystretch}{1.05}
\begin{tabular*}{\linewidth}{@{\extracolsep{\fill}}lcccc@{}}
\toprule Variant &\shortstack{Response\\supervision}&\shortstack{Feature\\update}&\shortstack{Temporal\\aggregation}&\shortstack{Motion\\bias}\\\midrule
Video-FT &--&--&--&--\\
Heads-only &Yes&--&--&--\\
Heads+Adapter &Yes&Yes&--&--\\
Full w/o bias &Yes&Yes&Yes&--\\
Full &Yes&Yes&Yes&Yes\\\bottomrule
\end{tabular*}
\par\endgroup

  \vspace{6pt}
  \noindent\begin{minipage}{\columnwidth}
\begingroup
  \centering
  \includegraphics[width=\linewidth]{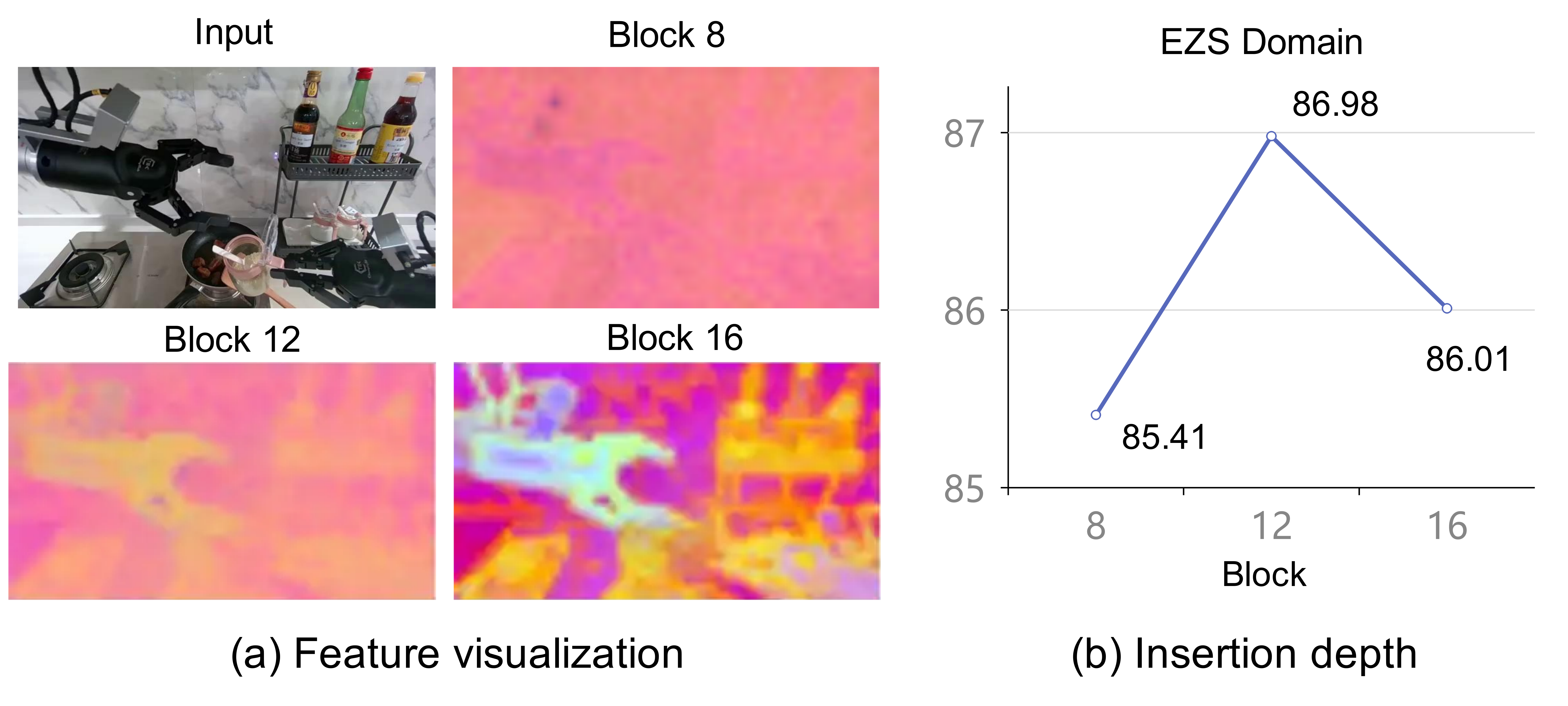}
  \par\captionsetup{type=figure,skip=4pt}
  \caption{\textbf{Insertion depth.}
    (a) Input image and PCA visualizations of backbone features at the same forward call and time position.
    (b) EZS-Bench Domain scores for different insertion blocks.}
  \label{fig:insertion}
  \par
\endgroup

  \end{minipage}\par
}
\section{Additional Ablations and Analysis}
\label{sec:ablation-config}
\newcommand{\RWAblationParagraph}[1]{\par\noindent\textbf{#1}\space}
Table~\ref{tab:ablation-config} lists the component variants.

\RWAblationParagraph{Whole-entity context.}
For $m_F$, we pool with the maximum first-frame interaction-role probability at each location in place of future response intensity.

\RWAblationParagraph{No zero reference.}
We use the conditioned adapter output as the object residual and omit the zero-dynamics subtraction.

\RWAblationParagraph{Amplitude matching.}
Scaling the No zero reference object residual to match the RMS magnitude of Full gives PSS 4.51 and Domain 85.95. Full scores 4.62 and 86.98.

\par\noindent\textbf{Insertion depth.}\space
Insertion at block~12 gives a Domain score of 86.98. Blocks~8 and~16 give 85.41 and 86.01 (Fig.~\ref{fig:insertion}).

\par\noindent\textbf{Bias-strength sensitivity.}\space
With Full fixed, we test bias multipliers of 0, 0.5, 1, 1.5, and 2. At zero, object feature updates and content attention are active.

\begingroup
\raggedright
\section{Additional Visualizations and Feedback Analysis}
\endgroup
\subsection{Additional Task Comparisons}
In Fig.~\ref{fig:more-comparison}(A), the position of the bottle opening relative to the receiving surface helps assess whether squeezing delivers jam to the intended target.
In (B), the comparison focuses on rim alignment during placement and support of the upper cup after release.

\subsection{Object Placement and Packing}
Fig.~\ref{fig:more-comparison-extra} focuses on the outcome of each transfer.
In (C), the pillow should rest on the sofa after release.
In (D), the carried item must pass through the bag opening and come to rest inside.

\begingroup
\raggedright
\subsection{Structured Responses and \texorpdfstring{\mbox{Motion-Guided Attention}}{Motion-Guided Attention}}
\endgroup
Fig.~\ref{fig:more-response} pairs predicted response fields with generated RGB frames.
For motion-guided attention, the feedback module adds a displacement bias to content logits within the fixed $7\times7$ candidate window.
The bias favors offsets near the predicted displacement, with its strength modulated by the motion gate.
Fig.~\ref{fig:more-attention} shows the bias and the resulting attention distribution.

\subsection{Directional Feedback}
We apply leftward and rightward offsets to the displacement input of the feedback module. The comparison in Fig.~\ref{fig:more-intervention} shows how the generated approach motion responds to each offset.
\par

\makeatletter
\if@firstcolumn
  \newpage
  \null\par
\fi
\makeatother
\clearpage
\endgroup

\twocolumn[{%
  \par\noindent
\begin{minipage}{\linewidth}%
  \includegraphics[width=\linewidth]{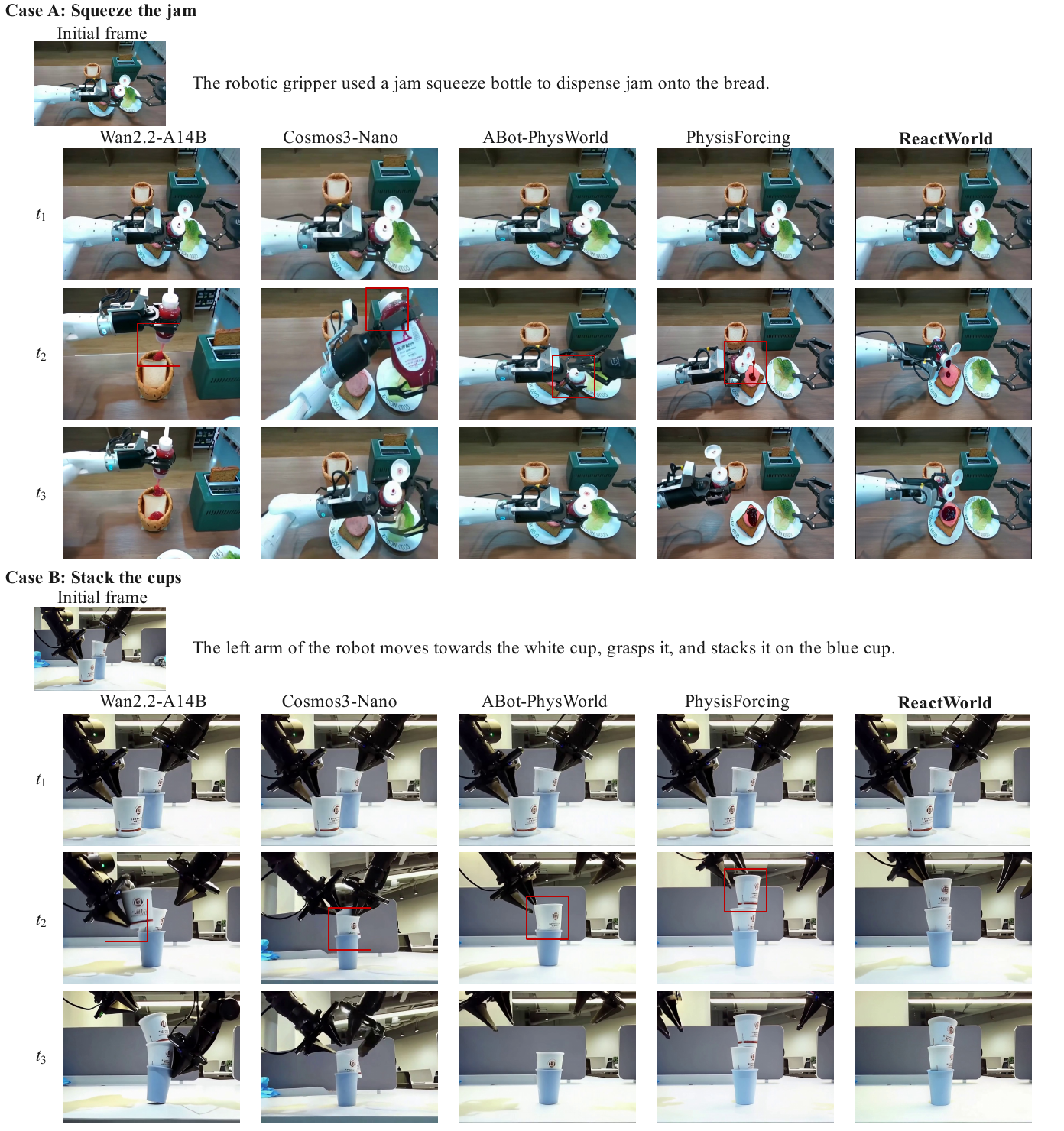}\par
  \FigureCaption{Qualitative comparisons for (A) squeezing jam and (B) stacking cups. For each task, we show the initial frame, task instruction, and three selected frames ($t_1$, $t_2$, and $t_3$) for Wan2.2-A14B, Cosmos3-Nano, ABot-PhysWorld, PhysisForcing, and ReactWorld.}{fig:more-comparison}%
\end{minipage}\par

}]
\null\par

\twocolumn[{%
  \par\noindent
\begin{minipage}{\linewidth}%
  \includegraphics[width=\linewidth]{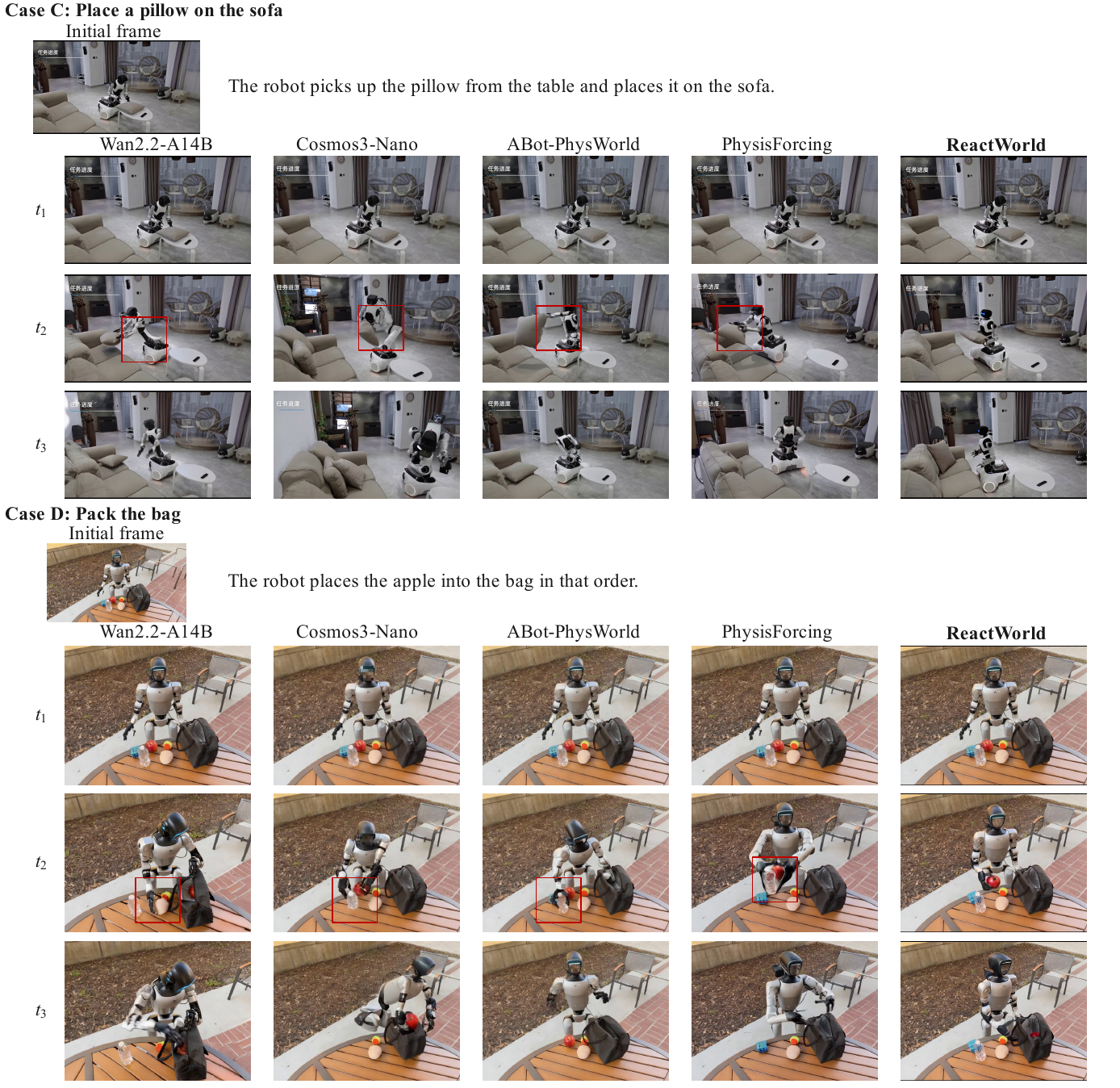}\par
  \FigureCaption{Qualitative comparisons for (C) placing a pillow on the sofa and (D) packing a bag. Each case shows the initial frame, task instruction, and three selected frames ($t_1$, $t_2$, and $t_3$) per method, following the five-method comparison in Fig.~\ref{fig:more-comparison}.}{fig:more-comparison-extra}%
\end{minipage}\par

}]
\null\par

\twocolumn[{%
  \par\noindent
\begin{minipage}{\linewidth}
\includegraphics[width=\linewidth]{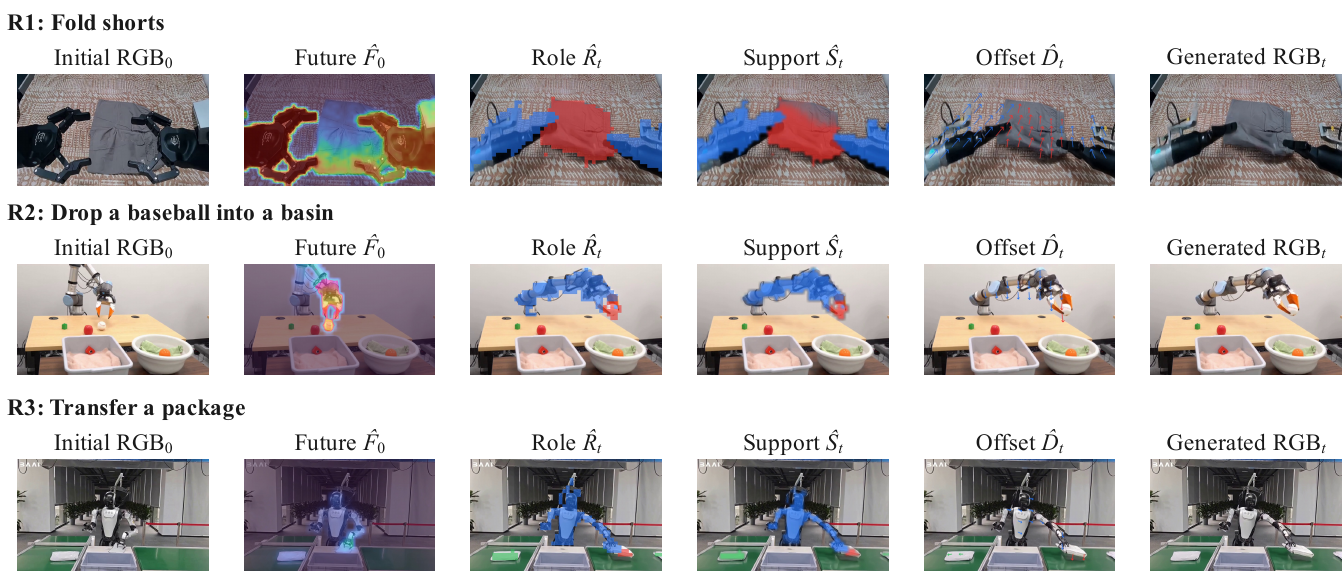}\par
\FigureCaption{Structured response fields and RGB frames for (R1) folding shorts, (R2) dropping a baseball into a basin, and (R3) transferring a package. The future response $F$ is referenced to the initial frame, $R$ describes participant roles at a state, and $S$ and $D$ describe motion support and displacement between adjacent states.}{fig:more-response}
\end{minipage}\par\smallskip

  \par\noindent
\begin{minipage}{\linewidth}

\includegraphics[width=\linewidth]{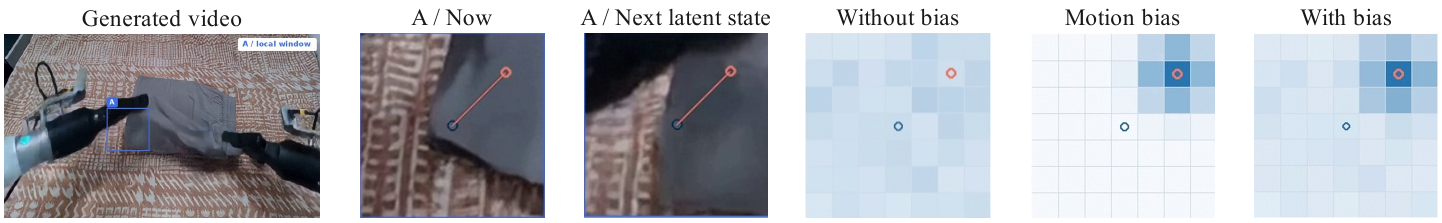}\par
\FigureCaption{Motion-guided attention for a fixed query and candidate window. The panels show content-only attention, the displacement bias, and attention with the bias. Both attention maps use the same query and key features.}{fig:more-attention}
\end{minipage}\par\smallskip

  \par\noindent
\begin{minipage}{\linewidth}
\includegraphics[width=\linewidth]{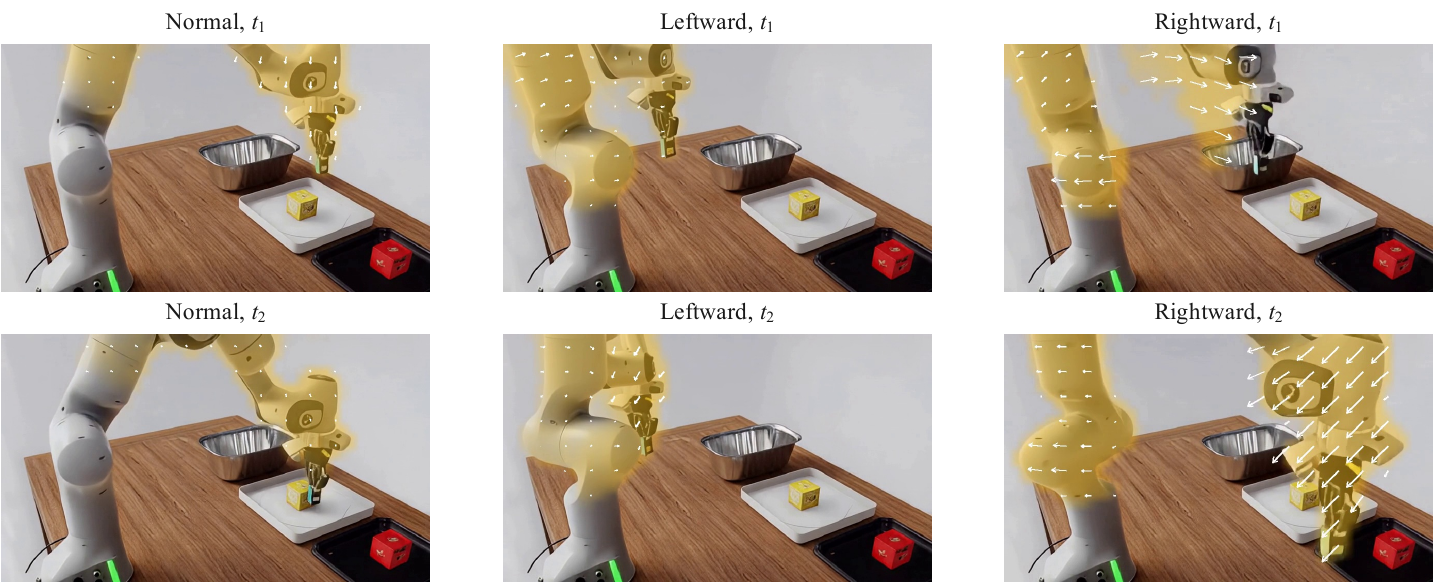}\par
\FigureCaption{Directional feedback for one input under normal feedback and leftward or rightward displacement offsets. The panels show two selected moments for each condition.}{fig:more-intervention}\end{minipage}\par\smallskip

}]
\clearpage
\enlargethispage{\dimexpr150pt-\textheight\relax}
{\small\bibliographystyle{supplement/ieeenat_fullname}\bibliography{supplement/references}}